\documentclass[letterpaper, 11 pt, conference]{ieeeconf}                         
\IEEEoverridecommandlockouts                             
\usepackage[utf8]{inputenc}
\newcommand{\cmark}{\ding{51}}
\newcommand{\xmark}{\ding{55}}
\usepackage{amssymb}
\usepackage{pifont}
\usepackage{lipsum}
\usepackage{graphicx}
\usepackage{pdflscape}
\usepackage{multicol,lipsum}
\usepackage{pgfplots}
\usepackage{graphicx}
\graphicspath{ {./images/} }
\usepackage{tabularx}
\usepackage{tablefootnote}
\usepackage{amsmath}
\usepackage{dblfloatfix}  
\pgfplotsset{compat = newest}
\usepackage{subcaption}
\usepackage{caption, babel}
\usepackage{gensymb}

\usepackage{pgfplots}
\usepackage[export]{adjustbox}
\usepackage{hyperref}
\title{\LARGE \bf
Fingerprint Pore Detection: A Survey
}
\usepackage{authblk}

\author{Azim Ibragimov, Mauricio Pamplona Segundo \\
Computer Science \& Engineering, University of South Florida \\
{\tt\small \{azim,mauriciop\}@usf.edu}}

\begin{document}

\maketitle
\thispagestyle{empty}
\pagestyle{empty}

\begin{abstract}
    This work presents the first survey on fingerprint pore detection. The survey provides a general overview of the field and discusses methods, datasets, and evaluation protocols. We also present a baseline method inspired on the state-of-the-art that implements a customizable Fully Convolutional Network, whose hyperparameters were tuned to achieve optimal pore detection rates. Finally, we also reimplementated three other approaches proposed in the literature for evaluation purposes. We have made the source code of  (1) the baseline method, (2) the reimplemented approaches, and (3) the training and evaluation processes for two different datasets available to the public to attract more researchers to the field and to facilitate future comparisons under the same conditions. The code is available in the following repository: \url{https://github.com/azimIbragimov/Fingerprint-Pore-Detection-A-Survey}
\end{abstract}

\section{INTRODUCTION}

Due to fingerprint recognition's adherence to the fundamental biometrics concepts of permanence and distinctiveness, it has been extensively researched during the past decade. This led to the development of numerous Automatic Fingerprint Identification Systems (AFIS). To distinguish between genuine and imposter individuals, the majority of AFIS use Level 1 (i.e., global characteristics - ridge flow and pattern type) and Level 2 (i.e., local features - minutiae and ridge skeleton) features \cite{Jain2007}. These techniques achieve high recognition accuracy but rely on high-quality, low resolution images to do so. \cite{segundo2015} Due to this, many researchers have become interested in using Level 3 features. They include all dimensional attributes of a ridge, such as ridge path deviation, width, shape, pores, edge contour, incipient ridges, breaks, creases, scars, and other permanent details. This wide range of fingerprint attributes made many researchers interested in using L3 features as supplemental data for matching \cite{Jain2007}. This statement is also supported by \cite{germannd} who claims that latent print examiners constantly consider L3 features while performing their duties since level 3 fingerprint features were included in the FBI standard. \cite{Jain2007}

Fingerprint sweat pores were proven to be perpetual, immutable, and unique \cite{locard1912}. The perpetuity, immutability, and uniqueness of fingerprint sweat pores make them a perfect fit for recognition purposes. \cite{germannd} 

To capture L3 features, one needs high-resolution, high-quality fingerprint images. The Scientific Working Group on Friction Ridge Analysis, Study, and Technology (SWGFAST) proposed a minimum scanning resolution of 1 000 dpi, which is considered to be high-resolution, for latent images \cite{swgfast06}. These high-resolution images can capture locations and shapes of fingerprint pores, which can be later used for recognition purposes. 

Sizes of fingerprint sweat pores range from 88 to 220 microns and they come in a variety of forms, including round, elliptical, oval, square, rhomboid, and triangular. Additionally, the location of the sweat pores can vary. They are typically located in the middle of a fingerprint ridge (closed pores), but they can sometimes occasionally be open to the side (open pores). \cite{locard1912} Figure 1 illustrates examples of sweat pores that are present in a high-resolution fingerprint image.

\begin{figure}[h]
    \centering
    \includegraphics[width=\columnwidth]{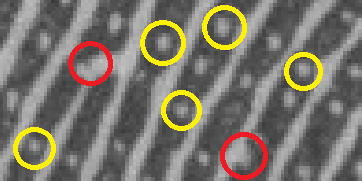}
    \caption{Sweat pores that are present in a high-resolution fingerprint image. Blue circles represent open pores, and yellow circles represents closed pores.}
    \label{fig:my_label}
\end{figure}

 This phenomena led to development of innovative AFIS systems that rely on pore detection. Since pores' shape and size can vary greatly, detecting them can be difficult \cite{stosz94, roddy97, swain14}. This difficulty inspired researchers to enhance pore detection results, and significant progress has been done so far \cite{labati18,ray2005, lemes2014}.

However, there are numerous issues associated with the publications that have been released in the past 25 years. We believe that addressing these problems would positively affect the quality of future publications. These issues include, but are not limited to, varying evaluation procedures, various training and testing protocols, difficulties in duplicating the results of other studies, and a lack of publications that make their source code publicly available. These factors lead to publications that are hard to analyze and compare. This publication aims to analyze these issues, offer solutions for some of them, and hopefully support future publications in this research field.

To our knowledge, no thorough survey on pore detection has ever been carried out. This paper aims to bring clarity to the subject, compare existing works, and figure out the best practices for fingerprint pore extraction

This study divides pore detection publications into two major categories: handcrafted approaches and machine learning approaches. Techniques developed by experts using various functions, transformations and filters to locate pores are referred to as handcrafted approaches. Techniques that use neural networks to locate pores are referred to as machine learning methods. For a while, handcrafted methods were preferred, but in recent years, there has been a surge in the number of publications using methods based on machine learning. Figure 2 displays the number of publications that have employed either handcrafted or machine learning approaches to locate pores.

\begin{figure}[h]
    \centering
    \begin{tikzpicture}
\begin{axis}[
    axis x line*=bottom,
    ymin=0,
     hide y axis,
    ybar stacked,
    bar width=12pt,
    nodes near coords,
    legend style={at={(0.5,-0.15)},
      anchor=north,legend columns=-1},
    symbolic x coords={'94,'05,'07,'08,'10,'12,'17,'18,'19, '21},
    xtick=data,
    ]
    
        \addplot+[ybar] plot coordinates {('94,1) ('05, 1)('07, 1)('08,1) ('10,2) ('12,1) ('17,0)('18,2)('19,0)('21,0)};
        \addplot+[ybar] plot coordinates {('94,0) ('05, 0)('07, 0)('08,0) ('10,0) ('12,0) ('17,4)('18,3)('19,2)('21,1)};

    \addplot+[ybar] plot coordinates {};

  \legend{\strut HC,\strut  ML}
  \end{axis}
\end{tikzpicture}
    \caption{Number of publications that have employed either handcrafted or machine learning approaches to locate pores. Blue bars represent number of publications that have employed handcrafted approaches and red bars represent number of publications that have employed machine learning approaches. }
    \label{fig:my_label}
\end{figure}
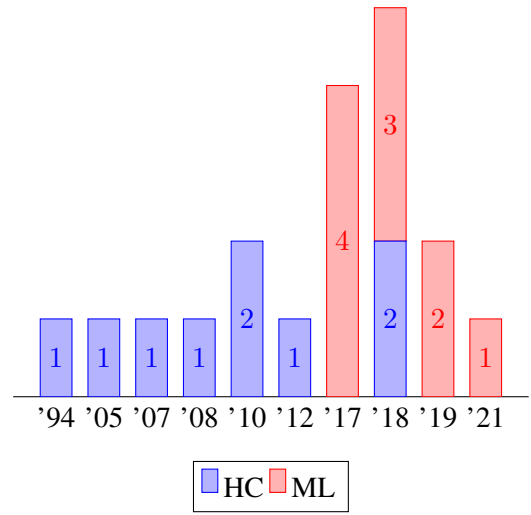

As a general rule, machine learning approaches tend to display better results when compared to handcrafted approaches. This can be seen when comparing various works that are either implement machine learning \cite{zhao2018, vijay2019, zuolin2019, vijay2020} or hand-crafted approaches \cite{jang17, ray2005}

Our contributions are:
\begin{itemize}
    \item The first survey on fingerprint pore detection, which covers existing materials and methods and summarizes the current state-of-the-art performance.
    \item A baseline approach inspired by state-of-the-art works, whose final configuration was decided based on extensive experiments.
    \item The reimplementation of three works from the literature~\cite{dahia2018,su2017,lemes2014}.
    \item A public repository containing the source code of reimplementations, baseline, and experiments to help future comparisons in this research field.
\end{itemize}

We organize this survey as follows. Section~\ref{sec:methods} introduces the different existing approaches divided into three main categories: handcrafted, machine learning, and hybrid approaches. Section~\ref{sec:datasets} presents the datasets used in the literature to report pore detection performance. Section~\ref{sec:baseline} introduces our proposed baseline and the layout definition process. Section~\ref{sec:evaluation} presents results reported in the literature and results obtained in this work for the proposed baseline and three reimplemented works. Finally, Section~\ref{sec:conclusion} presents our final remarks and discusses future directions.

\section{Pore detection methods}\label{sec:methods}
This paper covers several publications on pore detection. Ideally, they should follow these typical steps:
\begin{enumerate}
    \item Use a publicly available dataset of fingerprint images with annotated pore locations;
    \item Propose a new method or modifications to an existing one seeking to advance the literature on this topic;
    \item Employ a reproducible and fair protocol to validate the results of the new method;
    \item Compare the reported results with other publications, and discuss the advantages and shortcomings of the proposed method.
\end{enumerate}

In this section, we focus on the design aspects of each work. While most publications have unique approaches, it is possible to organize them into the following categories: handcrafted (skeleton-based and filtering-based), machine learning, and hybrid approaches. The following subsections present details about each of these categories.

\subsection{Handcrafted skeleton-based methods}
The early stages of pore detection research relied on skeleton tracking, a popular technique for fingerprint-related applications (e.g., ridge reconstruction \cite{malik2009} and minutia extraction \cite{farina1999}). With it, researchers showed that pore analysis could significantly improve fingerprint recognition results and sparked interest in the field \cite{stosz94, krzysztof2008}
    
\begin{figure}[h]
    \centering
    \begin{subfigure}[b]{0.32\columnwidth}
        \includegraphics[width=\columnwidth]{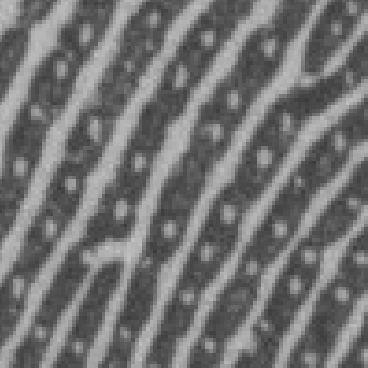}
        \caption{}
        \label{fig:skeletonized-a}
    \end{subfigure}
    \begin{subfigure}[b]{0.32\columnwidth}
        \includegraphics[width=\columnwidth]{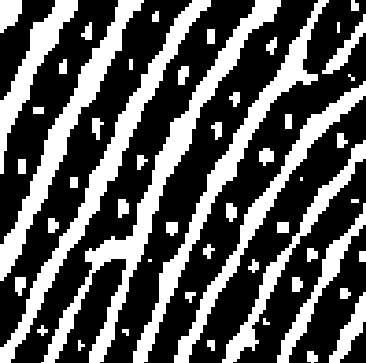}
        \caption{}
        \label{fig:skeletonized-b}
    \end{subfigure}
    \begin{subfigure}[b]{0.32\columnwidth}
        \includegraphics[width=\columnwidth]{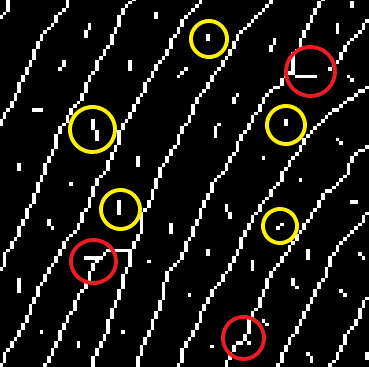}
        \caption{}
        \label{fig:skeletonized-c}
    \end{subfigure}
\caption{An example of (a) fingerprint image along its (b) binarized and (c) skeletonized versions with open (red circles) and closed pores (yellow circles).}
\label{fig:skeletonized}
\end{figure}

The core idea in this category of approaches is to binarize the fingerprint so that pores become white-pixel blobs and then to skeletonize the white areas so that pores show up as short-line segments. For instance, Stosz~and~Alyea~\cite{stosz94} navigate through these lines starting from endpoints (white pixels with a single white neighbor) and then stop if they reach another endpoint or a bifurcation within a maximum distance. Getting another endpoint means we have a closed pore, while a bifurcation indicates an open pore. Figure~\ref{fig:skeletonized} presents a skeletonized fingerprint image with closed and open pores highlighted in yellow and red circles, respectively.

Kryszczuk~\emph{et~al.}~\cite{krzysztof2008} exploited image preprocessing to improve the quality of the obtained skeleton. In this context, they used a Gabor wavelet filtering to enhance the ridges before skeletonization. Figure~\ref{fig:gaborFilterModel} illustrates the shape of the Gabor wavelet. As observed, this wavelet will produce strong responses for image locations containing bright and dark areas side-by-side, a common characteristic in fingerprint images. Figure~\ref{fig:hybrid-b} shows one example of the Gabor wavelet filtering outcome. This work spotlights one of the main weaknesses of this type of approach, which is its dependence on the skeleton quality to detect pores correctly. The next categories ease this problem by not relying on complicated leading stages.

\subsection{Handcrafted filtering-based methods}
Before the deep learning era, filtering-based strategies were the most common approach to detect pores~\cite{Jain2007,ray2005,zhao2008,mngenge2012,malathi2010,rodrigues2018,zaidi2018}. These works rely on human experts' ability to devise a sequence of filtering steps capable of highlighting pores in fingerprint images to a point where their detection becomes effortless. In the literature, popular strategies exploit the fact that fingerprints have evenly distributed pores along their friction ridges and that pores are small white blobs of pixels in fingerprint images.

\begin{figure}[h]
    \centering
    \begin{subfigure}[b]{0.49\columnwidth}
        \includegraphics[width=\columnwidth]{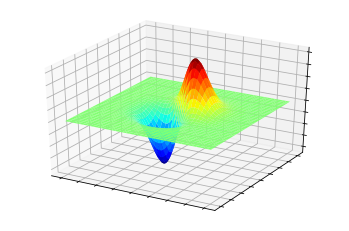}
        \caption{}
        \label{fig:gaborFilterModel}
    \end{subfigure}
    \begin{subfigure}[b]{0.49\columnwidth}
        \includegraphics[width=\columnwidth]{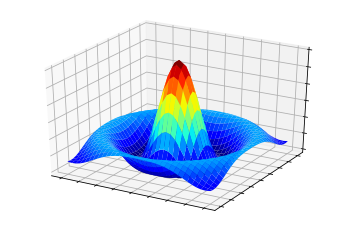}
        \caption{}
        \label{fig:mexicanHat}
    \end{subfigure}
    \begin{subfigure}[b]{0.49\columnwidth}
        \includegraphics[width=\columnwidth]{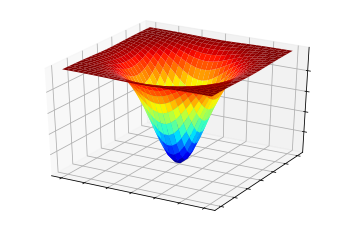}
        \caption{}
        \label{fig:raymodel}
    \end{subfigure}    
    \begin{subfigure}[b]{0.49\columnwidth}
        \includegraphics[width=\columnwidth]{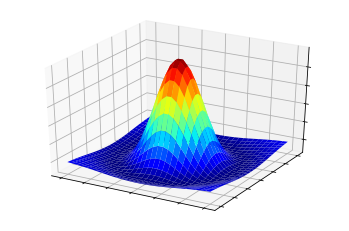}
        \caption{}
        \label{fig:parsonsmodel}
    \end{subfigure}    
    \begin{subfigure}[b]{0.49\columnwidth}
        \includegraphics[width=\columnwidth]{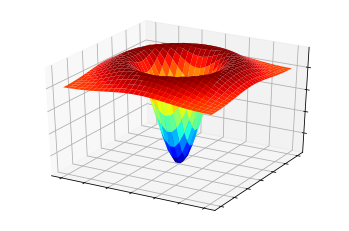}
        \caption{}
        \label{fig:parsonsmodel}
    \end{subfigure}
    \begin{subfigure}[b]{0.49\columnwidth}
        \includegraphics[width=\columnwidth]{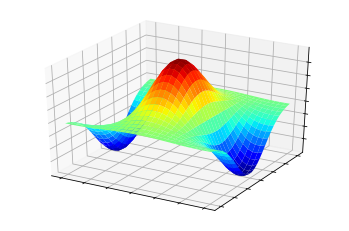}
        \caption{}
        \label{fig:dapm}
    \end{subfigure}  
    
    \caption{Illustrations of (a) a Gabor wavelet, commonly used to segment fingerprint ridges, and of (b) a Mexican Hat wavelet~\cite{Jain2007}, (c) a Gaussian kernel~\cite{ray2005}, (d) the Difference of Gaussians~\cite{parsons2007}, (e) the Laplace of Gaussian~\cite{mngenge2012}, and (f) an Anisotropic filter~\cite{zhao2008}, all employed to highlight pores.}
    \label{fig:filters}
\end{figure}

Jain~\emph{et~al.}~\cite{Jain2007} used Gabor and Mexican Hat wavelets to highlight ridges and pores, respectively. Figure~\ref{fig:mexicanHat} illustrates the shape of the Mexican Hat wavelet. As observed, it will produce strong responses for small bright areas surrounded by dark areas. This result, combined with the strong response in ridge areas produced by Gabor wavelets, will doubly reinforce the intensity of pixels in pore areas making the detection simply a matter of binary thresholding and blob localization. Other works also adopted isotropic kernels as pore models: Ray~\emph{et~al.}~\cite{ray2005} used a Gaussian kernel, Parsons~\emph{et~al.}~\cite{parsons2007} used the Difference of Gaussians, and Mngenge~\emph{et~al.}~\cite{mngenge2012} the Laplacian of Gaussian.

\begin{figure*}[!ht]
    \centering
    \includegraphics[width = \linewidth]{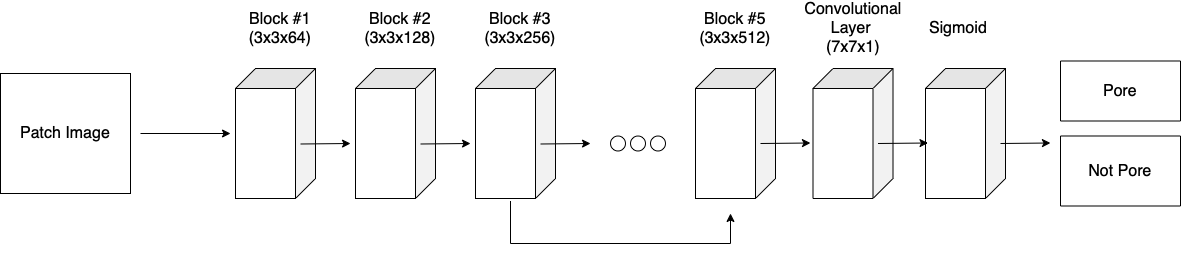}
    \caption{Diagram of the CNN architecture proposed by Shen~\emph{et~al.}~\cite{zuolin2019}.}
    \label{fig:su}
\end{figure*}

Zhao~\emph{et~al.}~\cite{zhao2008} presented a dynamic anisotropic pore model to handle open pores better since ridges do not surround them entirely (Figure~\ref{fig:dapm}). This model includes angle and scale parameters to adapt to common pore variations. They estimate these parameters from local ridge orientation and frequency. Rodrigues~and~Borges~\cite{rodrigues2018} built upon this work by using anisotropic filters instead of Gaussians for a DoG-based detection.

Finally, some works invested in an image processing style of solution. Malathi~\emph{et~al.}~\cite{malathi2010} used watershed transform to locate ``water-filled'' pores. Lemes~\emph{et~al.}~\cite{lemes2014} counted the number of transitions between white and black pixels in the border of a size-adjusted circle around a pixel to decide whether it is a pore or not.

However, researchers found this group of techniques computationally expensive and hard to adapt to unseen imaging conditions (\emph{e.g.}, image resolution, quality, and finger pressure). For this reason, they started exploring machine learning methods to overcome these issues.

\subsection{Machine Learning approaches}\label{sec:ml}
Unlike handcrafted methods that rely on experts'  knowledge to devise techniques to locate pores, machine learning leverages annotated data to model pores automatically. This approach type received lots of attention after deep neural networks became popular. A neural network is a learning system loosely inspired by the brain. It consists of layers of interconnected neurons that find relevant patterns within the training data. Within the pore detection field, a specific type of neural network - Convolutional Neural Network (CNN)~\cite{LeCun1989} - is a unanimous choice. One reason is CNN's ability to take advantage of data with spatial contexts, such as images. The convolutional layers can learn several filters to extract the most significant visual features for a particular problem.

\begin{figure}[!ht]
    \begin{subfigure}[h]{\columnwidth}
        \centering
        \includegraphics[width=0.18\linewidth]{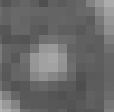}
        \includegraphics[width=0.18\linewidth]{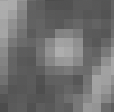}
        \includegraphics[width=0.18\linewidth]{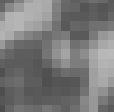}
        \includegraphics[width=0.18\linewidth]{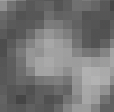}
        \includegraphics[width=0.18\linewidth]{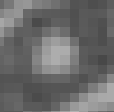}
        \caption{Positive patch samples}
        \label{fig:pospatches}
    \end{subfigure}

    \begin{subfigure}[h]{\columnwidth}
        \centering
        \includegraphics[width=0.18\linewidth]{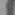}
        \includegraphics[width=0.18\linewidth]{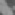}
        \includegraphics[width=0.18\linewidth]{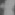}
        \includegraphics[width=0.18\linewidth]{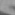}
        \includegraphics[width=0.18\linewidth]{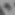}
        \caption{Negative patch samples}
        \label{fig:negPatch}
    \end{subfigure}

    \caption{Patches with (a) positive cases and (b) negative cases that are fed into CNN}
    \label{fig:patches}
\end{figure}

\begin{figure}[!ht]
    \centering
    \begin{subfigure}[h]{0.49\columnwidth}
         \includegraphics[width=\linewidth]{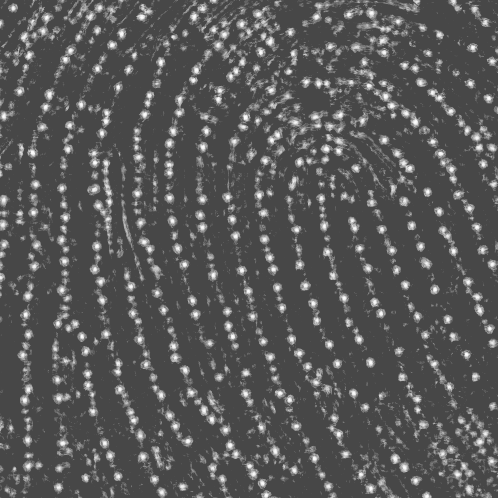}
          \caption{}
          \label{fig:beforeNMS}
    \end{subfigure}
    \begin{subfigure}[h]{0.49\columnwidth}
        \includegraphics[width=\linewidth]{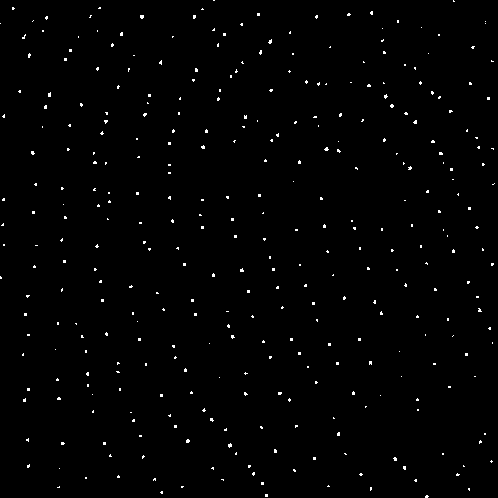}
        \caption{}
        \label{fig:afterNMS}
    \end{subfigure}
    \caption{(a) Pore Intensity Map and (b) Binary Pore Map}
    \label{fig:patches}
\end{figure}

Shen~\emph{et~al.}~\cite{zuolin2019} used the Fully Convolutional Network (FCN)~\cite{Long2015} architecture displayed in Figure \ref{fig:su} for pore detection. Their architecture has five blocks with two convolutional and one max-pooling layer each, and a residual connection between the third and fifth blocks. Convolutional layers are followed by ReLU activation~\cite{Glorot2011} (except the last one) and max-pooling layers by batch normalization~\cite{Ioffe2015}. A final convolutional layer projects the output into a 1-dimensional space with sigmoid activation. This network takes 17x17 image patches as input, and the outcome indicates whether the input patch has a pore on its center or not. Figure~\ref{fig:pospatches} displays samples of patches centered on pores, and Figure~\ref{fig:negPatch} patches not centered on pores. They train the network with focal loss to overcome data unbalancing issues since the number of non-pore patches is much higher than the number of pore patches. For inference, the FCN generates a pore intensity map with the probability of each pixel being the center of a pore, as shown in Figure~\ref{fig:beforeNMS}. The non-maximum suppression (NMS) post-processing algorithm~\cite{hosang2017} merges neighboring pixels with high probability values into a single detection to obtain unique coordinates for each pore. The final result is a binary pore map, as shown in Figure~\ref{fig:afterNMS}, with discrete points representing the detected pores.

Other works presented similar approaches \cite{jang17,vijay2019,zhao2018,su2017,wang2017,ding2021,dahia2018,labati18} with minor modifications when compared to Shen~\emph{et~al.}'s work. Jang~\emph{et~al.}~\cite{jang17} employed soft labels to the training patches to ease the transition between pore and non-pore labels using a Gaussian distribution. Anand~\emph{et~al.}~\cite{vijay2019} exploited residual networks~\cite{He2016} to facilitate the training of deep models. Labati~\emph{et~al.}~\cite{labati18} included a second CNN to refine pore detection results and remove possible errors from the pore map. Finally, Wang~\emph{et~al.}~\cite{wang2017} and Ding~\emph{et~al.}~\cite{ding2021} used a U-Net~\cite{Ronneberger2015} architecture instead of relying on a patch-based classification.

\subsection{Hybrid Approaches}

Hybrid approaches combine handcrafted and machine learning techniques in a way that overcomes the limitations of each strategy. This way, they can rely on human ingenuity to overcome training data limitations and ease the learning task.
          
\begin{figure}[h]
    \centering
    \begin{subfigure}[h]{0.32\columnwidth}
        \includegraphics[width=\linewidth]{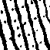}
        \caption{}
        \label{fig:hybrid-a}
    \end{subfigure}
    \begin{subfigure}[h]{0.32\columnwidth}
        \includegraphics[width=\linewidth]{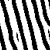}
        \caption{}
        \label{fig:hybrid-b}
    \end{subfigure}
    \begin{subfigure}[h]{0.32\columnwidth}
        \includegraphics[width=\linewidth]{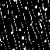}
        \caption{}
        \label{fig:hybrid-c}
    \end{subfigure}

    \begin{subfigure}[h]{0.32\columnwidth}
        \includegraphics[width=\linewidth]{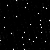}
        \caption{}
        \label{fig:hybrid-d}
    \end{subfigure}
    \begin{subfigure}[h]{0.32\columnwidth}
        \includegraphics[width=\linewidth]{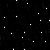}
        \caption{}
        \label{fig:hybrid-e}
    \end{subfigure}
    \caption{ (a) Binary pore map, (b) ridge map, (c) Results of applying XOR operation on binary and ridge maps, (d) Pore Intensity Map, (e) Binary Pore Map.}
\end{figure}

Zhao~\emph{et~al.}~\cite{zhao2018} devised a filter-based image enhancement to facilitate pore detection using a CNN. Like in other filtering-based approaches, they binarize the fingerprint (Figure~\ref{fig:hybrid-a}) and segment ridges using Gabor filters (Figure~\ref{fig:hybrid-b}) and locate white pixels in both images with an XOR operation to be used as pore candidates (Figure~\ref{fig:hybrid-c}). The result becomes the input of a CNN that finds actual pores in the input image (Figure~\ref{fig:hybrid-d}). The post-processing stage also uses an XOR operation between the CNN output and the ridges map to remove false detections (Figure~\ref{fig:hybrid-e}).

\section{Datasets}\label{sec:datasets}
For publications on pore detection, datasets are essential. They must include high-resolution fingerprint images and annotated details about pores and their locations in this field of study. To our knowledge, eight datasets have been used in the past by pore detection publications. Table~\ref{tab:datasets} provides detailed information about these datasets. 

\begin{table}[h]
    \centering

     \begin{tabularx}{0.45\textwidth} { 
      | >{\centering\arraybackslash}X 
      | >{\centering\arraybackslash}X 
      | >{\centering\arraybackslash}X 
      | >{\centering\arraybackslash}X 
      | >{\centering\arraybackslash}X 
      |>{\centering\arraybackslash}X | }    
      \hline
        \small Ref. & \small Name & \small dpi & \small Size & \small Public? & \small Year \\ \hline \hline
        \small\cite{NIST} & \small NIST SD 4 & \small 500 & \small 10 & \small No* & \small 2005 \\ \hline
        \small\cite{zhao2008} &\small PolyU-HRF &\small 1200 &\small 30 (DBI), 120 (DBII) & \small No* & \small 2010 \\ \hline
        \small\cite{genovese2016} &\small GEN &\small 3800 &\small 66 & \small No & \small 2016 \\ \hline
        \small\cite{zhao2018} &\small DB Touch-less, Latent &\small 1000 &\small 80 & \small Yes & \small 2018 \\ \hline 
        \small\cite{vijay2022} &\small IITI HRF &\small 1000 &\small 20 & \small No & \small 2021 \\ \hline
        \small\cite{ding2021} &\small Sub-cutan-eous &\small 2000 &\small 60 & \small No  & \small 2021  \\ \hline
        \small\cite{lemes2014} &\small NB\_ID &\small 1000 &\small 30 & \small No  & \small 2014  \\ \hline
        \small\cite{wyzykowski2020} &\small L3-SF &\small 1200 &\small 740 & \small Yes  & \small 2020 \\  \hline
    \end{tabularx}
    \caption{Summary of datasets that are used by publications in pore extraction. No* refers to datasets that were publicly available in the past, but not anymore }
    \label{tab:datasets}
\end{table}

Researchers must follow an evaluation protocol for each dataset. This protocol outlines how to use the dataset for experimentation. Earlier publications that proposed handcrafted methods used datasets for testing purposes only. They developed their algorithms and evaluated their performance on the entire dataset. More recent papers that use machine learning techniques divide the datasets into training, validation, and testing subsets. These subsets serve as a means for model learning, hyperparameter tuning, and performance evaluation. This section will refer to the datasets from Table~\ref{tab:datasets} and outline how pore detection publications used them.

\subsection{NIST SD 4 (withdrawn)}
One of the earliest works on pore identification used the NIST Special Dataset 4 (NIST SD 4)~\cite{ray2005}. To our knowledge, this is the only publication to use this dataset, which is no longer distributed due to the lack of documentation required by NIST~\cite{NIST}. The dataset contains thousands of 500 dpi fingerprints, and Ray~\emph{et~al.}~\cite{ray2005} annotated pore locations of 10 cherry-picked images for their pore detection experiments. 

\subsection{PolyU HREF (withdrawn)}
The following dataset, the Hong-Kong Polytechnic University High-Resolution Fingerprint Database (PolyU-HRF) \cite{zhao2008}, was used by numerous publications~\cite{zhao2008, ray2005, teixeira2014, rodrigues2018, labati18, su2017, wang2017, jang17, dahia2018, zuolin2019, ali2021}. It consists of 3170 images with 1200 dpi and has two pore-annotated subsets. The first subset, DBI, is composed of partial fingerprints of size $320\times240$. This set contains 30 images with annotated pore locations. The second subset, DBII, has full-size fingerprint images with $640\times480$ pixels. This subset has 120 images with pores annotated by Teixeira~and~Leite~\cite{teixeira2017}. Unfortunately, this dataset is no longer distributed by the authors.

\subsection{IITI-HRF (public)}
The Indian Institute of Technology Indore High-Resolution Fingerprint Database (IITI-HRF)~\cite{vijay2022} is a more recent dataset and consists of two subsets. The first one, IITI-HRFC, contains 6400 images with 1000x1000 pixels at 1000 dpi. The second, IITI-HRFP, is a 320x240 center crop of the first subset. The authors provide pore annotations for 20 images of the IITI-HRFP subset~\cite{vijay2022}. This is currently the only publicly available dataset of high-resolution fingerprints.

\subsection{L3-SF (public)}
L3 synthetic fingerprint (L3-SF) database~\cite{wyzykowski2020} is a publicly available dataset that contains synthetic high-resolution fingerprint images with the same structure as the PolyU-HRF DBI subset. The dataset also contains a subset of 740 images with 1200 dpi and annotations of pore locations. As it is composed of synthetic fingerprints, L3-SF avoids privacy-related issues for dataset distribution.

\subsection{Private datasets}

Different works created their own private datasets for pore detection analysis, but these were not used by other works later on. The private datasets are:

\subsubsection{GEN (private)}
Genovese~\emph{et~al.}~\cite{genovese2016} created a dataset to develop a pore extraction technique for touchless fingerprint images. Touchless fingerprint images do not require fingers to be placed on sensors, thus allowing for the prevention of hygiene problems and distortions~\cite{genovese2016}. Their dataset consists of $1500\times1000$ that are centered on fingerprint core. A dataset comprised 66 touchless fingerprint pictures.

\subsubsection{DB Touch-less, DB Latent (private)}
Labati~\emph{et~al.}~\cite{labati18} created two datasets, DB Touch-less and DB Latent, with 80 annotated images in total. These images have resolution of at least 1000 dpi and were used to evaluate pore detection in heterogeneous conditions.

\subsubsection{Subcutaneous (private)}
The datasets discussed in this publication so far have only utilized 2D images. Optical coherence tomography was used by Ding~\emph{et~al.}~\cite{ding2021} to gather data about 0–3 mm beneath the skin's surface. It contains 60 pore-annotated images with $1500\times1200$ pixels at 2000 dpi.

\subsubsection{NB\_ID (private)}
They created the NB\_ID dataset to address the problem of newborn identification. It contains three subsets, NB\_ID\_A, NB\_ID\_B, and NB\_ID\_II, with thousands of newborn palmprints and footprints. For their experiments, Lemes~\emph{et~al.}~\cite{lemes2014} randomly selected ten images per subset and annotated pores in a 100x100 patch from each of them. 

\section{CNN Baseline specification}\label{sec:baseline}

\begin{figure*}
    \centering
    \includegraphics[width = 0.9\linewidth]{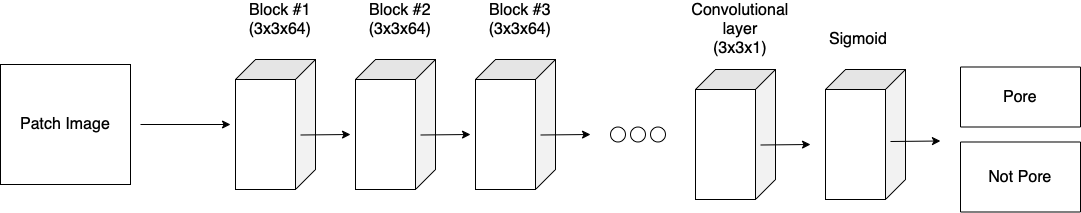}
    \caption{Baseline CNN Architecture}
    \label{fig:baseline}
\end{figure*}

Reproducing the results of other papers is difficult, as none of them released their source code. To address this problem, we release every pore detection method implemented in our publication with instructions on how to replicate the results of our experiments. 
Our first implemented approach is a CNN baseline inspired by existing machine learning approaches (Section~\ref{sec:ml}). We designed a customizable FCN, and Figure~\ref{fig:baseline} illustrates its architecture. It consists of applying $3\times3$ convolutions with valid padding to the input patch until the outcome is a $1\times1$ value representing the probability of the patch being centered on a pore. Thus, the number of layers depends on the input patch size. This network can be modified in the following ways:

\begin{description}
    \item[\bf Patch size:]\hfill\\Input patches must fully enclose pores. A small size would not capture the entire pore, while a large size would bring too much unnecessary information. Recent publications used patch size of $17\times17$~\cite{zuolin2019,su2017,dahia2018,wang2017} and $19x19$~\cite{vijay2022} pixels. In our experiments, we considered patch sizes of $13\times13$, $15\times15$, $17\times17$, and $19\times19$ pixels.
    \item[\bf Pore radius:]\hfill\\In the labelling process, all pixels inside a circle with radius $r$ centered on the pore are labeled with $1$, and other pixels receive the label $0$. This radius must be be able to represent shape of a pore. Recent publications used $r$ values of $3$~\cite{su2017,dahia2018} and $5$~\cite{vijay2022}. In our experiments, we considered $r$ values of $4$, $5$, and $6$.
    \item[\bf Number of trainable layers:]\hfill\\The depth of the network determines its learning capacity. But the more layers we use, the more trainable parameters we have. Some publications used max-pooling layers~\cite{zuolin2019,wang2017} to reduce the dimensionality of intermediary tensors without requiring any trainable parameter. We considered architectures with and without max-pooling layers in our experiments. When using max-pooling, we replace every second convolutional layer with a max-pooling with the same kernel size, thus obtaining an architecture with half of the trainable layers.
    \item[\bf Residual connections:]\hfill\\Residual connections add skip connections between layers or blocks of layers to speed up the training and avoid the vanishing gradients problem in deep networks. Recent publications used residual networks for pore detection~\cite{vijay2022,zuolin2019}. We evaluated architectures with and without residual connections for internal blocks (all except the first and last blocks).
    \item[\bf Soft labels:]\hfill\\In hard labeling, pixels receive label $0$ or $1$. Meanwhile, soft labeling assigns a value in the range $[0,1]$ proportional to the distance between the pixel and the center of the closest pore. Su~\emph{et~al.}~\cite{su2017} utilized soft labels in their approach. In our approach, we labeled pixels inside a circle with radius $r$ around a pore with the value $(r-d)/r$, and other pixels with $0$. Our experiments evaluate training with hard and soft labels.
\end{description}

    \begin{table}
     \centering
     \begin{tabularx}{0.48\textwidth} { 
      | >{\centering\arraybackslash}X 
      | >{\centering\arraybackslash}X 
      | >{\centering\arraybackslash}X 
      | >{\centering\arraybackslash}X 
      | >{\centering\arraybackslash}X 
      | >{\centering\arraybackslash}X 
      |>{\centering\arraybackslash}X | }     
      \hline
        \small Patch size & \small Pore Radius & \small Max Pooling & Res. connec-tions & DBI I F-score & DBII II F-score \\ \hline

        \small 13 & \small 4 &  \small \xmark & \small \xmark & 89.45 &  90.61\\ \hline
        \small 13 & \small 5 &  \small \xmark & \small \xmark & 88.67 & 90.40\\ \hline
        \small 13 & \small 6 &  \small \xmark & \small \xmark & 89.03 & 90.50\\ \hline

        \small 13 & \small 4 &  \small \cmark & \small \xmark & 82.85 & 84.91\\ \hline
        \small 13 & \small 5 &  \small \cmark & \small \xmark & 85.20 & 85.74\\ \hline
        \small 13 & \small 6 &  \small \cmark & \small \xmark & 85.41 & 87.15\\ \hline
        
        \small 13 & \small 4 &  \small \xmark & \small \cmark & 87.77 & 89.78\\ \hline
        \small 13 & \small 5 &  \small \xmark & \small \cmark & 87.77 & 89.82\\ \hline
        \small 13 & \small 6 &  \small \xmark & \small \cmark & 88.00 & 89.87\\ \hline

        \small 13 & \small 4 &  \small \cmark & \small \cmark & 82.67 & 84.23\\ \hline
        \small 13 & \small 5 &  \small \cmark & \small \cmark & 85.62 & 87.87\\ \hline
        \small 13 & \small 6 &  \small \cmark & \small \cmark & 86.24 & 86.80\\ \hline

        \small 15 & \small 4 &  \small \xmark & \small \xmark & 90.62 & 91.11\\ \hline
        \small 15 & \small 5 &  \small \xmark & \small \xmark & 89.69 & 91.44\\ \hline
        \small 15 & \small 6 &  \small \xmark & \small \xmark & 89.70 & 91.26\\ \hline

        \small 15 & \small 4 &  \small \cmark & \small \xmark & 88.22 & 87.74\\ \hline
        \small 15 & \small 5 &  \small \cmark & \small \xmark & 88.01 & 89.58\\ \hline
        \small 15 & \small 6 &  \small \cmark & \small \xmark & 87.87 & 89.69\\ \hline
        
        \small 15 & \small 4 &  \small \xmark & \small \cmark & 88.88 & 90.28\\ \hline
        \small 15 & \small 5 &  \small \xmark & \small \cmark & 87.65 & 90.00\\ \hline
        \small 15 & \small 6 &  \small \xmark & \small \cmark & 88.00 & 90.25\\ \hline

        \small 15 & \small 4 &  \small \cmark & \small \cmark & 87.65 & 89.22\\ \hline
        \small 15 & \small 5 &  \small \cmark & \small \cmark & 88.09 & 89.76\\ \hline
        \small 15 & \small 6 &  \small \cmark & \small \cmark & 87.79 & 89.44\\ \hline

        \small 17 & \small 4 &  \small \xmark & \small \xmark & 89.79 & 90.64\\ \hline
        \small 17 & \small 5 &  \small \xmark & \small \xmark & 90.90 & 92.24\\ \hline
        \small 17 & \small 6 &  \small \xmark & \small \xmark & 90.45 & 91.40\\ \hline

        \small 17 & \small 4 &  \small \cmark & \small \xmark & 87.66 & 86.56\\ \hline
        \small 17 & \small 5 &  \small \cmark & \small \xmark & 88.41 & 86.67\\ \hline
        \small 17 & \small 6 &  \small \cmark & \small \xmark & 88.41 & 84.33\\ \hline

        \small 17 & \small 4 &  \small \xmark & \small \cmark & 87.91 & 89.77\\ \hline
        \small 17 & \small 5 &  \small \xmark & \small \cmark & 87.49 & 89.90\\ \hline
        \small 17 & \small 6 &  \small \xmark & \small \cmark & 87.91 & 90.05\\ \hline

        \small 17 & \small 4 &  \small \cmark & \small \cmark & 83.19 & 86.33\\ \hline
        \small 17 & \small 5 &  \small \cmark & \small \cmark & 85.39 & 87.73\\ \hline
        \small 17 & \small 6 &  \small \cmark & \small \cmark & 82.70 & 84.26\\ \hline

        \small 19 & \small 4 &  \small \xmark & \small \xmark & 91.11 & 92.09\\ \hline
        \small 19 & \small 5 &  \small \xmark & \small \xmark & 90.91 & 91.92\\ \hline
        \small 19 & \small 6 &  \small \xmark & \small \xmark & 90.01 & 91.39\\ \hline

        \small 19 & \small 4 &  \small \cmark & \small \xmark & 89.67 & 90.47\\ \hline
        \small 19 & \small 5 &  \small \cmark & \small \xmark & 89.09 & 90.81\\ \hline
        \small 19 & \small 6 &  \small \cmark & \small \xmark & 88.64 & 90.45\\ \hline

        \small 19 & \small 4 &  \small \cmark & \small \cmark & 88.82 & 90.16\\ \hline
        \small 19 & \small 5 &  \small \cmark & \small \cmark & 88.61 & 90.39\\ \hline
        \small 19 & \small 6 &  \small \cmark & \small \cmark & 87.49 & 90.11\\ \hline

        \small 19 & \small 4 &  \small \xmark & \small \cmark & 87.12 & 89.43\\ \hline
        \small 19 & \small 5 &  \small \xmark & \small \cmark & 87.88 & 89.49\\ \hline
        \small 19 & \small 6 &  \small \xmark & \small \cmark & 88.63 & 90.70\\ \hline

    \end{tabularx}
    \caption{Experiment results on PolyU-HRF dataset for different patch sizes and pore radii, with or without max-pooling, and with or without residual connections. All these experiments use soft labels (see Table~\ref{table:experiments}).}
    \label{table:experimentsPolyU}
\end{table}

    \begin{table}
     \centering
     \begin{tabularx}{0.48\textwidth} { 
      | >{\centering\arraybackslash}X 
      | >{\centering\arraybackslash}X  
      |>{\centering\arraybackslash}X | }
      \hline
        \small Soft labels & DBI I F-score & \small DBII II F-score \\ \hline 
        \small \xmark & 88.32 & 90.5 \\ \hline
        \small \cmark & 90.90 & 92.24 \\ \hline
    \end{tabularx}
    \caption{Effects of using hard or soft labels when using $17\times17$ input patches, pore radius equal to $5$, and not including max-pooling layers or residual connections.}
    \label{table:experiments}
\end{table}

\subsection{Training details}

Aside from the previously mentioned parameters, all other parameters were identical for all experiments to prevent other factors from affecting the performance results. We use a batch size of $128$ patches, the Adam optimizer with a learning rate of $0.001$, early stopping when the training does not improve on validation for $10$ consecutive epochs. We investigate the best combination of parameters using the PolyU-HRF dataset. We use 120 annotated images from DBII~\cite{teixeira2017}, which are split into training (90), validation (5) and testing (25), and we also use 30 images from DBI for testing.

\subsection{Suggested configuration}

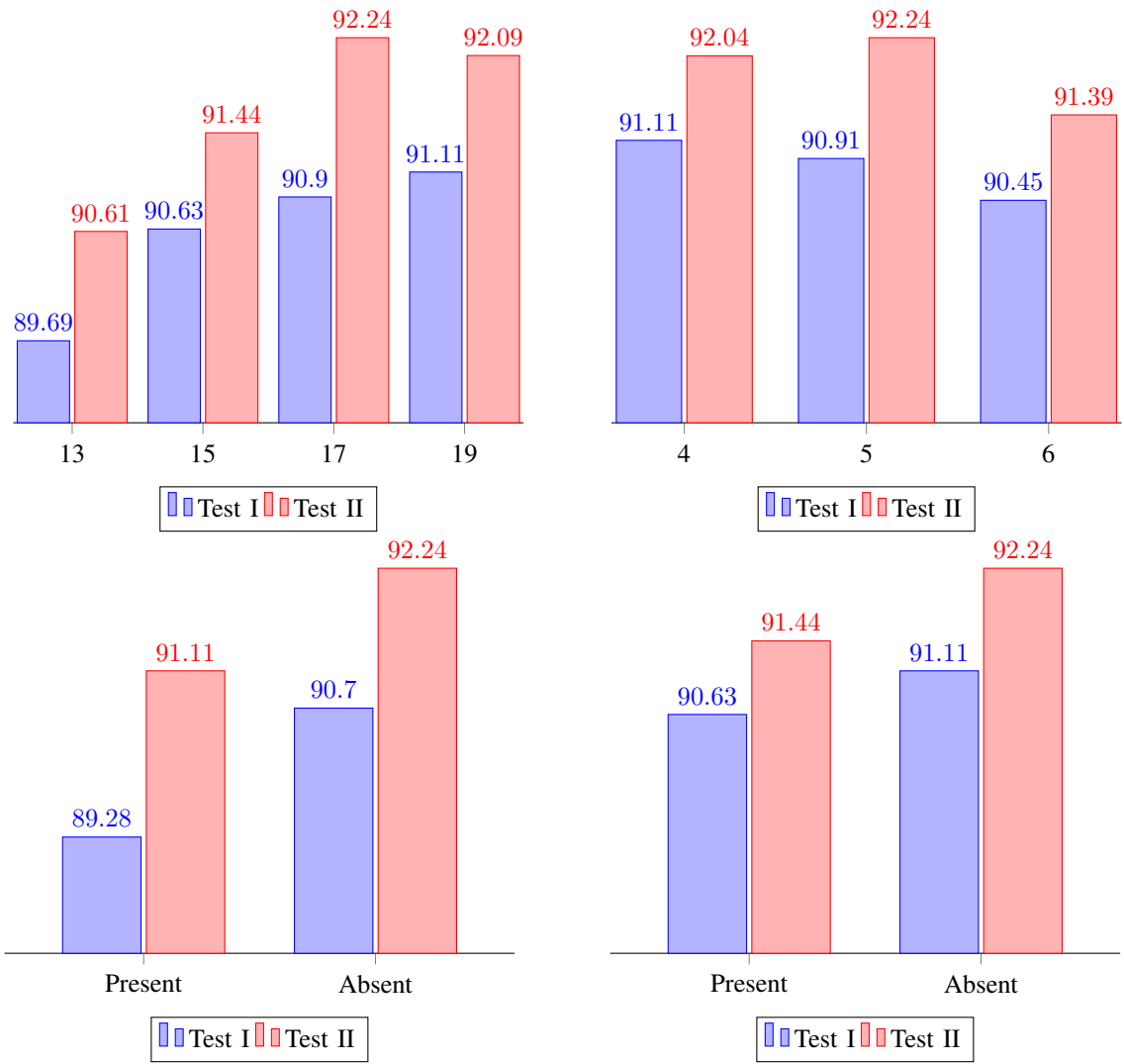
\begin{figure*}
        \centering
        \begin{subfigure}{0.45\textwidth}
             \begin{tikzpicture}
                    \tikzset{
                      font={\fontsize{10pt}{12}\selectfont}}
                    \begin{axis}[
                        axis x line*=bottom,
                        ymin=89,
                        hide y axis,
                        ybar,
                        bar width=20pt,
                        enlarge x limits=0.15,
                        legend style={at={(0.5,-0.15)},
                          anchor=north,legend columns=-1,},
                        symbolic x coords={13,15,17,19},
                        xtick=data,
                        nodes near coords, 
                        ]

                    \addplot coordinates {(13, 89.69) (15, 90.63) (17, 90.90) (19, 91.11)};
                    \addplot coordinates {(13, 90.61) (15, 91.44) (17, 92.24) (19, 92.09)};
                    
                    \legend{Test I,Test II}
                    \end{axis}
                    \end{tikzpicture}
        \end{subfigure}
        \begin{subfigure}{0.45\textwidth}
             \begin{tikzpicture}
                    \tikzset{
                      font={\fontsize{10pt}{12}\selectfont}}
                    \begin{axis}[
                        axis x line*=bottom,
                        ymin=88,
                        hide y axis,
                        ybar,
                        bar width=25pt,
                        enlarge x limits=0.2,
                        legend style={at={(0.5,-0.15)},
                          anchor=north,legend columns=-1,},
                        symbolic x coords={4,5,6},
                        xtick=data,
                        nodes near coords, 
                        ]

                    \addplot coordinates {(4, 91.11) (5, 90.91) (6, 90.45)};
                    \addplot coordinates {(4, 92.04) (5, 92.24) (6, 91.39)};
                    
                    \legend{Test I,Test II}
                    \end{axis}
                    \end{tikzpicture}

            \end{subfigure}
            
        \begin{subfigure}{0.45\textwidth}
             \begin{tikzpicture}
                    \tikzset{
                      font={\fontsize{10pt}{12}\selectfont}}
                    \begin{axis}[
                        axis x line*=bottom,
                        ymin=88,
                        hide y axis,
                        ybar,
                        bar width=30pt,
                        enlarge x limits=0.6,
                        legend style={at={(0.5,-0.15)},
                          anchor=north,legend columns=-1,},
                        symbolic x coords={Present,Absent},
                        xtick=data,
                        x tick label style={ align=center},
                        nodes near coords, 
                        ]
                    \addplot coordinates {(Present, 89.28) (Absent, 90.70)};
                    \addplot coordinates {(Present, 91.11) (Absent, 92.24)};
                    \legend{Test I,Test II}
                    \end{axis}
                    \end{tikzpicture}

            \end{subfigure}
            \begin{subfigure}{0.45\textwidth}
                \begin{tikzpicture}
                    \tikzset{
                      font={\fontsize{10pt}{12}\selectfont}}
                    \begin{axis}[
                        axis x line*=bottom,
                        ymin=88,
                        hide y axis,
                        ybar,
                        bar width=30pt,
                        enlarge x limits=0.6,
                        legend style={at={(0.5,-0.15)},
                          anchor=north,legend columns=-1,},
                        symbolic x coords={Present,Absent},
                        xtick=data,
                        x tick label style={ align=center},
                        nodes near coords, 
                        ]
                    \addplot coordinates {(Present, 90.63) (Absent, 91.11)};
                    \addplot coordinates {(Present, 91.44) (Absent, 92.24)};
                    \legend{Test I,Test II}
                    \end{axis}
                    \end{tikzpicture}
            \end{subfigure}

        \caption{ Affect of (a) patch sizes, (b) pore radius, (c) max pooling, and (d) residual connections on detection results.}
        \label{fig:experimentstatistics}
    \end{figure*}

We have conducted a grid search over the customizable parameters of our baseline to assess their impact on pore detection performance. Each combination of parameters was repeated three times to reduce the effects of the random initialization, and we report the average results. The results for these experiments are shown in Tables~\ref{table:experimentsPolyU}~and~\ref{table:experiments}, and a graphical comparison is shown in Figure~\ref{fig:experimentstatistics}. After analyzing the results, we ended up with the following model configuration: 

\begin{itemize}
    \item Patch sizes of $17\times17$ and $19\times19$ produce the best results. These patch sizes are large enough to capture various shapes and sizes of pores. We used $17\times17$ patches since their models are more resource efficient and produce results comparable to $19\times19$.
    \item Pore radius of $4$ and $5$ produce the best results for the PolyU-HRP images with 1200 dpi. We used a pore radius of $5$ since it provided better results for a $17\times17$ patch. 
    \item Not using max-pooling layers produces better results, meaning that using more layers produce more accurate results. It is known that the expressive power of neural networks grow exponentially with the number of layers~\cite{Raghu2017}, so having more convolutional layers increased the ability of our baseline distinguishing between pore and non-pore patches.
    \item The use of residual connections decreased the performance. We attribute this result to the fact that residual networks perform better with very deep networks, while our CNN models are shallow.
    \item Soft labels produced better results than hard labels, as we do not have neighboring patches with opposite labels.
\end{itemize}

\section{Performance evaluation}\label{sec:evaluation}

\subsection{Correct detection criteria}

One major issue in pore detection publications and their respective evaluation protocols is the lack of a standard criterion to determine if a pore detection is correct. Nowadays, these publications use different criteria to report their performance, making the comparison of different works a complicated task. Table~\ref{tab:criteria} summarizes the evaluation criteria of existing publications. In this work, we adopt the bidirectional correspondence (\emph{i.e.}, a pore detection is correct if it is the closest detection to its closest ground truth pore), as this is the only criterion that does not accepts multiple detections for a single ground truth pore and also does not require any distance threshold.

\begin{table}[!ht]
     \centering
     \begin{tabularx}{0.48\textwidth} { 
      | p{\dimexpr.20\linewidth-2\tabcolsep-1.3333\arrayrulewidth}
      |>{\centering\arraybackslash}X | }     
      \hline
        \small Ref & \small Criteria  \\ \hline \hline
        \small\cite{genovese2016} \cite{ali2021}\cite{teixeira2014} \cite{vijay2020}  & Euclidean distance with empirical threshold
        \\ \hline
        \small\cite{labati18, zuolin2019, dahia2018, ding2021}  & Bidirectional correspondence    
        \\ \hline
        \small\cite{labati18}\cite{vijay2019} \cite{zhao2018} & Euclidean distance with threshold proportional to the average ridge width
        \\ \hline

        \small\cite{rodrigues2018}  & Manhattan distance with empirical threshold
        \\ \hline
    \end{tabularx}
    \caption{Criteria used by existing publication to determine if a pore detection is correct.}
    \label{tab:criteria}
\end{table}

\begin{table}
     \centering
     \begin{tabularx}{0.48\textwidth} { 
      | >{\centering\arraybackslash}X 
      | >{\centering\arraybackslash}X 
      | >{\centering\arraybackslash}X 
      | >{\centering\arraybackslash}X 
      | >{\centering\arraybackslash}X 
      |>{\centering\arraybackslash}X | }    
      \hline
        \small Protocol  & \small Ref  & \small  Training & \small Validation & \small Testing\\ \hline

        \small I & \small\cite{zhao2008, ray2005, Jain2007}  & \small 0 & \small 0 & \small 24 (DBI)  \\ \cline{1-5}
        \small II & \small \cite{teixeira2014, rodrigues2018, Jain2007, ray2005, zhao2008, su2017}  &  \small 0 & \small 0 & \small 30 (DBI)  \\ \cline{1-5}
        \small III & \small\cite{su2017} &  \small 20 (DBI) & \small 0 & \small 10 (DBI)  \\ \cline{1-5}
        \small IV & \small\cite{jang17, zhao2018}&  \small 18(DBI) & \small 6(DBI) & \small 6(DBI)  \\ \cline{1-5}
        \small V & \small\cite{ali2021} &  \small 24 (DBI) & 0 & \small 6 (DBI) \\ \hline
        \small VI & \small\cite{dahia2018} &  \small 15 (DBI) & \small 5 (DBI) & \small 10 (DBI)  \\ \cline{1-5}
        \small VII & \small\cite{wang2017} &  \small 70 (DBII) & \small 0 & \small 30 (DBI)  \\ \cline{1-5}
        \small VIII & \small \cite{zuolin2019, vijay2019, jang17, dahia2018} &  \small 90 (DBII) & 0 & \small 30 (DBI), 30 (DBII) \\ \cline{1-5}

    \end{tabularx}
    \caption{Data split protocols employed in publications using PolyU-HRF.}
    \label{tab:protocol}
\end{table}

\subsection{Literature evaluation using PolyU-HRF}

Although the PolyU-HRF dataset is no longer publicly available, it was heavily used in the literature of pore analysis. But even when using the same dataset, different works use different evaluation protocols (\emph{e.g.}, data splits and correct detection criteria) that hamper a direct comparison between them. Table~\ref{tab:protocol} lists all evaluation protocols used for PolyU-HRF and all publications that used the same protocol. We can observe that early handcrafted approaches used protocols that require no training or validation data. Meanwhile, machine learning publications used many different protocols, the most popular being Protocol VIII. Protocol VIII benefits from having an extensive training set and two testing subsets with slightly different conditions (partial (DBI) vs. full-size fingerprints (DBII)).

Table~\ref{tab:polyu-results} shows the results of publications that have used the PolyU-HRF dataset. We have included an F-score column reporting the harmonic mean of the true detection and false detection rates reported by the listed works. Even though so many different protocols were used, we can observe that machine learning approaches (Protocols III to VIII) clearly outperform handcrafted approaches (Protocols I and II). This is an expected outcome, as a similar behavior was observed in many different computer vision research fields. However, picking the best machine learning approach is not as obvious. If we group the protocols by the subset of data used for training, we get Protocols III to VI for DBI and Protocols VII and VIII for DBII. In the first group, the best performance was obtained by Ali~\emph{et~al.}~\cite{ali2021}, and, in the second group, by Shen~\emph{et~al.}~\cite{zuolin2019}. Both works use FCNs with some residual connections, with Ali~\emph{et~al.}~\cite{ali2021} using a shallower network combined with a handcrafted post-processing step. Although there is no clear winner between the two, Protocol VIII is more well founded than than Protocol V, making the results reported by Shen~\emph{et~al.}~\cite{zuolin2019} more reliable. Finally, we can see that the performance of our baseline is comparable to the top-performing works for the two most popular protocols (VI and VIII).

\begin{table}[!ht]
     \begin{tabularx}{0.5\textwidth} { 
      | >{\centering\arraybackslash}X 
      | >{\centering\arraybackslash}X 
      | >{\centering\arraybackslash}X 
      | >{\centering\arraybackslash}X 
      |>{\centering\arraybackslash}X | } 
      \hline
        \small Protocol & \small Ref & \small True Detection & \small False Detection & \small F-score   \\ \hline \hline
        \small I & \small\cite{zhao2008}  &  \small 84.8 & 17.6 & 83.58  \\  
        \cline{2-5} 
        \small  & \small\cite{ray2005}  & \small 60.6 & 30.5 & 64.74\\ 
                \cline{2-5} 
        \small  & \small\cite{Jain2007}  & \small 75.9 & 23.0 & 76.44 \\ \hline 
        \hline
        
        \small II & \small\cite{teixeira2014}  & \small 86.1 & 8.6 & 88.67 \\         
        \cline{2-5}
        \small  & \small\cite{Jain2007}  & \small 75.9 & 23.0 & 76.44 \\ 

        \cline{2-5}
        \small  & \small\cite{zhao2008}  & \small 84.8 & 17.6 & 83.58 \\

        \cline{2-5}
        \small  & \small\cite{rodrigues2018}  & \small 83.1 & 14.9 &  84.08 \\

        \cline{2-5}
        \small  & \small\cite{ray2005}  & \small 63.4 & 20.4 &  70.58 \\
        
        \cline{2-5}
        \small  & \small\cite{lemes2014}  & \small 90.8 & 11.1 & 89.83 \\

        \hline \hline 
        
        \small III & \small\cite{su2017}  & \small 88.6 & 0.4 & 93.77\\ \hline \hline

                \small IV & \small\cite{jang17}  & \small 93.09 &  8.64 & 92.21 \\
                \cline{2-5}
        \small  & \small\cite{zhao2018}  & \small 93.14 & 4.39 & 94.35 \\ \hline \hline

        \small V & \small\cite{ali2021}  & \small 96.69 & 4.18 & 96.24 \\
        \hline \hline

        \small VI & \small\cite{dahia2018}  & \small 91.95 & 8.88 & 91.53 \\
        \cline{2-5}
        \small  & \small\cite{su2017}  & \small 90.31 & 23.99 & 82.54 \small  \\
        \cline{2-5}
        \small  & \small Baseline  & \small 92.87 & 8.43 & 92.21 \small  \\
        \hline \hline

        \small VII & \small\cite{wang2017}  & \small 83.65  & 13.89 & 84.86
        \\ \hline \hline

        \small VIII & \small\cite{zuolin2019}  & \small 95.56 (DBII), 92.81 (DBI) & 8.1 (DBII), 7.3 (DBI) &  93.69 (DBII), 92.75 (DBI)\\
        \cline{2-5}
        \small  & \small\cite{vijay2019}  & \small 94.49 (DBII), 93.78 (DBI) & 8.5 (DBII), 8.5 (DBI) & 92.97 (DBII), 92.62 (DBI) \\
        \cline{2-5}
        \small  & \small\cite{jang17}  & \small 91.32 (DBII), 89.54 (DBI) & 8.5 (DBII), 8.5 (DBI) & 91.40 (DBII), 89.50 (DBI) \\
        \cline{2-5}
        \small  & \small\cite{dahia2018} &  \small 95.14 (DBII), 91.13 (DBI)  & 17.82 (DBII), 16.52 (DBII) & 88.18 (DBI) 87.13 \\
        \cline{2-5}
        \small  & \small\cite{su2017} &  \small 88.79 (DBII), 82.56 (DBI)  &  14.49 (DBII), 12.61 (DBII) & 87.12 (DBII) 84.90 (DBI) \\
        \cline{2-5}
        \small  & \small Baseline  &  93.74 (DBII), 91.79 (DBI) & 9.21 (DBII) 9.96 (DBI) & 92.24 (DBII) 90.90 (DBI) \\
        \hline

    \end{tabularx}
    \caption{Protocol table of publications that utilized PolyU Dataset}
    \label{tab:polyu-results}
\end{table}

\subsection{Literature evaluation using L3-SF}

Since the PolyU-HRF dataset is no longer distributed by the authors, we ran experiments using the L3-SF dataset so that researchers without access to the PolyU dataset can compare their results to our baseline approach. We also included our reimplementation of three literature approaches: Lemes~\emph{et~al.}~\cite{lemes2014}, Dahia~\emph{et~al.}~\cite{dahia2018}, and Su~\emph{et~al.}~\cite{su2017}. These reimplementations obtained results on PolyU-HRF that are comparable to the original works when using the same protocol, and we made their code publicly available as well. Since L3-SF contains 740 images, we divided them into five subsets and ran 5-fold cross validation. Table~\ref{tab:l3sf-experiments} shows the obtained results for the four considered approaches. As observed, machine learning approaches obtained a higher performance, although Dahia~\emph{et~al.}'s did not obtain consistent results for this dataset.

\begin{table}
     \centering
     \begin{tabularx}{0.48\textwidth} { 
      | >{\centering\arraybackslash}X 
      |>{\centering\arraybackslash}X | }
      \hline
        \small Name & \small F-score \\ \hline 
        \small Baseline & \small 87.20 \\ \hline
        \small DPF \cite{lemes2014} & \small 71.91 \\ \hline
        \small CNN1 \cite{su2017} & \small 86.86 \\ \hline
        \small CNN2 \cite{dahia2018} & \small 58.18 \\ \hline
    \end{tabularx}
    \caption{Baseline results on L3-SF}
    \label{tab:l3sf-experiments}
\end{table}

\subsection{Discussion and conclusion}\label{sec:conclusion}

This survey reviewed methods, datasets, and evaluation protocols used in pore detection works. It provides an overview of the existing techniques and how they compare to each other in terms of accuracy. Additionally, we presented a baseline approach that reflects the core ideas of state-of-the-art publications and made its source code publicly available. We also reimplemented three other works from the literature and added their code to our public repository. We hope to help future comparisons in this research field by releasing our code for pore detection methods and performance evaluation.

Although different approaches in the literature present high accuracy, the performance in recent years seems to have stagnated. Since the performance in existing datasets is not saturated, it would be interesting to investigate the performance of human annotators in this task to verify the precision of the ground truth annotations and their impact on the performance analysis of pore detectors.

For future directions, many handcrafted and machine learning techniques could be adapted to the problem of detecting pores and have never been tried before. For instance, researchers could use multitask learning to estimate local orientation and frequency, segment ridges, and detect pores and minutia simultaneously. By including additional fingerprint-related tasks and forcing the network to learn weights that are useful for different purposes, the performance of pore detection itself could increase. Furthermore, graph neural networks could be employed to explore pixel connectivity patterns for pore detection or as a post-processing technique that learns the distribution of pores in a fingerprint. Finally, all existing points deal with pores as 2D points, mainly due to the limited annotations available. But pores have different shapes and sizes. Researchers could explore this information not only to obtain more accurate pore detectors but also to provide additional information for pore-based fingerprint matching.

\bibliographystyle{IEEEtran}
\bibliography{main}
\end{document}